\documentclass[letterpaper, preprint, paper,11pt]{AAS}		

\usepackage{bm}
\usepackage{amsmath,amssymb,amsfonts}
\usepackage{subfigure}
\usepackage{gensymb}
\usepackage{todonotes}
\usepackage[colorlinks=true, pdfstartview=FitV, linkcolor=black, citecolor= black, urlcolor= black]{hyperref}
\usepackage{overcite}
\usepackage{footnpag}			      	

\PaperNumber{21-739}

\begin{document}

\title{NASA/GSFC’s Flight Software Core Flight System Implementation For A Lunar Surface Imaging Mission}

\author{Mohammed Eleffendi\thanks{Masters Student, Electrical Engineering and Computer Science, Embry-Riddle Aeronautical University, 1 Aerospace Blvd, Daytona Beach, FL., 32114},  
Daniel Posada\thanks{Ph.D. Candidate, Aerospace Engineering, Embry-Riddle Aeronautical University, 1 Aerospace Blvd, Daytona Beach, FL., 32114},
M. \.{I}lhan Akba\c{s}\thanks{Assistant Professor, Electrical Engineering and Computer Science, Embry-Riddle Aeronautical University, 1 Aerospace Blvd, Daytona Beach, FL., 32114},
\ and Troy Henderson\thanks{Assistant Professor, Aerospace Engineering, Embry-Riddle Aeronautical University, 1 Aerospace Blvd, Daytona Beach, FL., 32114}
}

\maketitle{} 		
\begin{abstract}
The interest in returning to the Moon for research and exploration has increased as new tipping point technologies are providing the possibility to do so.
One of these initiatives is the Artemis program by NASA, which plans to return humans by 2024 to the lunar surface and study water deposits in the surface.
This program will also serve as a practice run to plan the logistics of sending humans to explore Mars.
To return humans safely to the Moon, multiple technological advances and diverse knowledge about the nature of the lunar surface are needed.
This paper will discuss the design and implementation of the flight software of EagleCam, a CubeSat camera system based on the free open-source core Flight System (cFS) architecture developed by NASA's Goddard Space Flight Center.
EagleCam is a payload transported to the Moon by the Commercial Lunar Payload Services Nova-C lander developed by Intuitive Machines.
The camera system will capture the first third-person view of a spacecraft performing a Moon landing and collect other scientific data such as plume interaction with the surface.
The complete system is composed of the CubeSat and the deployer that will eject it.
This will be the first time WiFi protocol is used on the Moon to establish a local communication network.

\end{abstract}

\section{Introduction}\label{sec: Introduction}
In 2019, NASA's Commercial Lunar Services (CLPS) contracts allowed the research and development of multiple, rapid, commercial deliveries of research instruments and technology demonstrations from private and governmental parties to the lunar surface.
The first CLPS contracts were awarded to Intuitive Machines, Astrobotic, and Orbit Beyond\cite{Daines_2019}. With this award Intuitive Machines plans on taking a variety of scientific payloads to the lunar surface using the Nova-C lunar lander, providing a valuable approach to gather scientific data.

Intuitive Machines challenged Embry-Riddle Aeronautical University (ERAU) students to design, build, test, and operate a deployable camera to capture images of the landing of their lander, Nova-C.
EagleCam was developed for this purpose, and it will provide a full 360$^{\circ}$ field-of-view (FOV) of the spacecraft landing and landing site by using an array of cameras.

This paper focuses on the flight software development of EagleCam and the deployer that will eject it. A modular and reusable flight software that could be used in a later similar mission is desired.
NASA has developed open-source frameworks for developing modular and reusable flight software. One of these is the F Prime (F') spaceflight software released by NASA's Jet Propulsion Laboratory which is a component-driven framework and enables rapid development and deployment of spaceflight and other embedded software applications.\cite{bocchino2018f}
There is also the core Flight System (cFS) released by NASA's Goddard Space Flight Center which is a platform and project independent reusable software framework and set of reusable software applications.\cite{prokop2014nasa}
Both frameworks satisfy the modularity and reusability requirements and have flight heritage. However, cFS was chosen for this mission because it is easier to interface with the Nova-C flight software.


The second section of the paper will focus on briefly describing the mission and its requirements, the third section will provide a brief introduction about the cFS and discuss the flight software developed by ERAU, the fourth section will discuss the hardware part of the system, the fifth section will discuss several tests that the system underwent, and finally some conclusions about the cFS.

\section{Mission Background}
\begin{figure}[h!]
    \centering
    \includegraphics[width=0.8\textwidth]{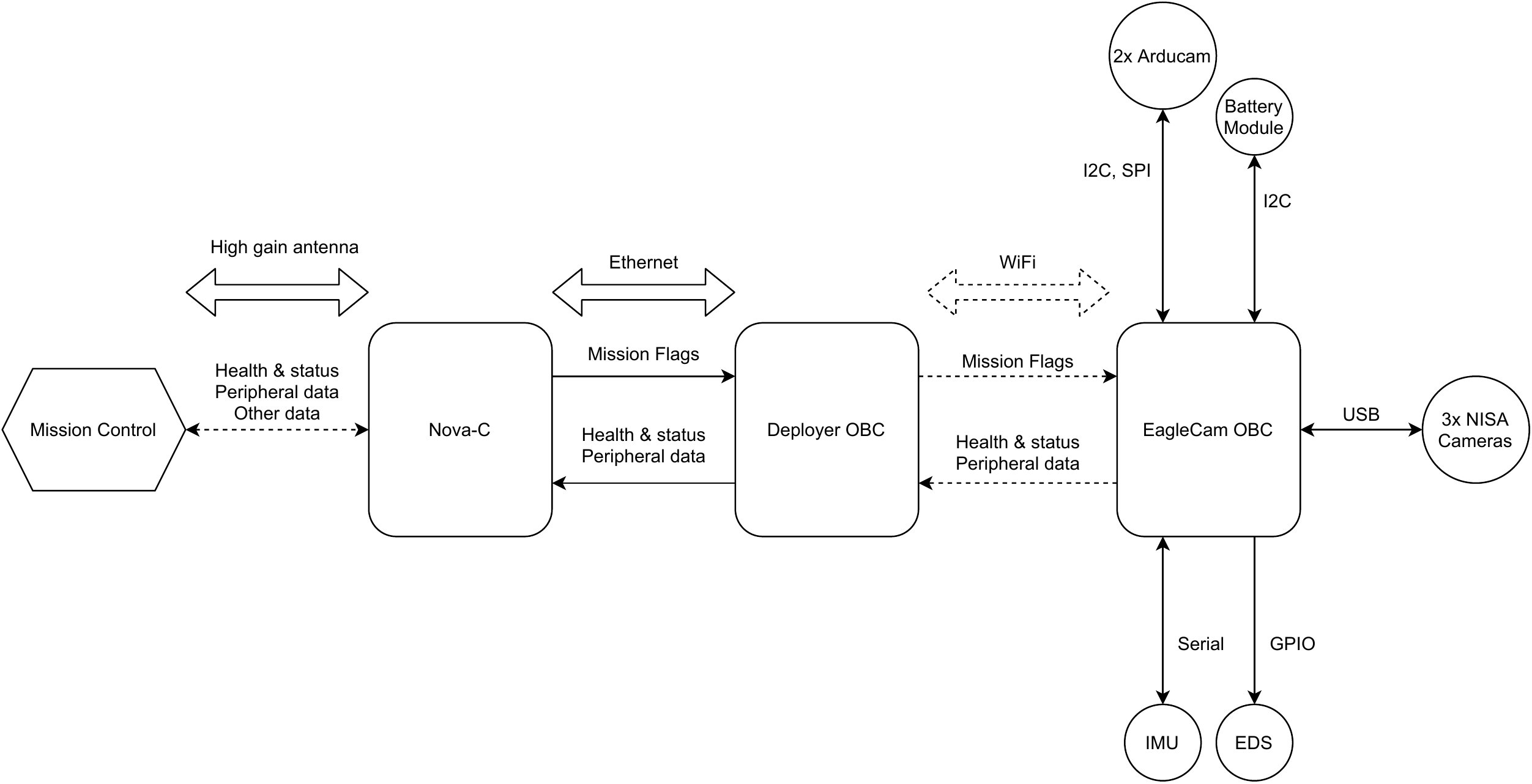}
    \caption{EagleCam and deployer subsystems.}
    \label{fig:system}
\end{figure}

EagleCam is a 1.5U CubeSat that will take a third person view of a landing on the lunar surface.
In order to achieve this, the system utilizes two on-board computers (OBCs) and a few peripherals as seen in Fig. \ref{fig:system}.
One OBC is inside the deployer, and a second inside the CubeSat.
The deployer OBC is connected to the EagleCam OBC via WiFi.
It is also connected to the Nova-C lander through Ethernet.
Nova-C will transmit the data it receives from the deployer OBC to mission control through its high gain antenna.

The system developed by ERAU
consists of the EagleCam CubeSat and the deployer that will deploy it to the lunar surface. The CubeSat consists of the EagleCam OBC and the following peripherals:

\begin{enumerate}
    \item Inertial Measurement Unit (IMU) which is a 9 DOF sensor module that will transmit the EagleCam acceleration, gyroscope and magnetometer measurements to the EagleCam OBC through a serial connection starting just before deployment and ending after EagleCam has landed on the lunar surface.
    \item Three Nano Immersive Situational Awareness (NISA) cameras, each one with a 186$\degree$ FOV lens, which will take pictures of the Nova-C lunar lander as it is landing on the lunar surface.
    Each NISA camera contains an internal OBC that is connected to the EagleCam OBC via USB and will run its own software upon receiving a trigger from the EagleCam OBC.
    All NISA cameras' OBCs are running the same software that will continuously take pictures and store them on the NISA OBC in which it is running.
    The EagleCam OBC will download these pictures from each NISA OBC using the File Transfer Protocol (FTP).
    \item Two ArduCam camera sensors that will take pictures of Nova-C before and after the Electrodynamic Dust Shield (EDS) is activated. 
    The main function of these two cameras is to test the EDS efficacy (The experiment will be described more in the following sections).
    Each ArduCam is connected to the EagleCam OBC via I2C which is used to initialize the camera, and SPI which is used for image transmission to the EagleCam OBC.
    \item An EDS that will be mounted in front of the ArduCam lenses.
    It will be controlled by the flight software using a GPIO pin that turns the EDS on and off. The main function of the EDS is to clean the camera lenses from the lunar dust that will accumulate on them after landing.
    \item A battery module that houses the Lithium-Ion batteries which will power on the EagleCam OBC and its peripherals.
    This battery module has an I2C interface to read the battery status.
\end{enumerate}

EagleCam is contained within the deployer during launch and flight.
It will be under transit lunar injection for approximately six days before approaching lunar orbit.
Then Nova-C will perform a few orbits for navigation purposes, and finally it will start the descent maneuver. 

When Nova-C reaches lunar orbit and performs terrain relative navigation, it will start the descent operations to navigate to the intended landing site.
EagleCam will be deployed to the lunar surface at approximately 30 $m$ from the surface when the lander is on the terminal descent phase where there is only vertical velocity.

The EagleCam OBC will transmit health and status messages in addition to all telemetry collected by its peripherals to the deployer OBC via WiFi.
The deployer OBC stores the data collected by the EagleCam peripherals which will later be sent to Nova-C, and forwards health and status messages from the EagleCam OBC to Nova-C which sends them to mission control. In addition, the deployer OBC forwards the necessary mission flags to the EagleCam OBC upon receiving them from Nova-C. All communication between the deployer OBC and Nova-C is done via Ethernet.

\begin{figure}[h!]
        \centering
        \begin{minipage}[]{0.8\textwidth}
            \centering
            \includegraphics[width=\textwidth]{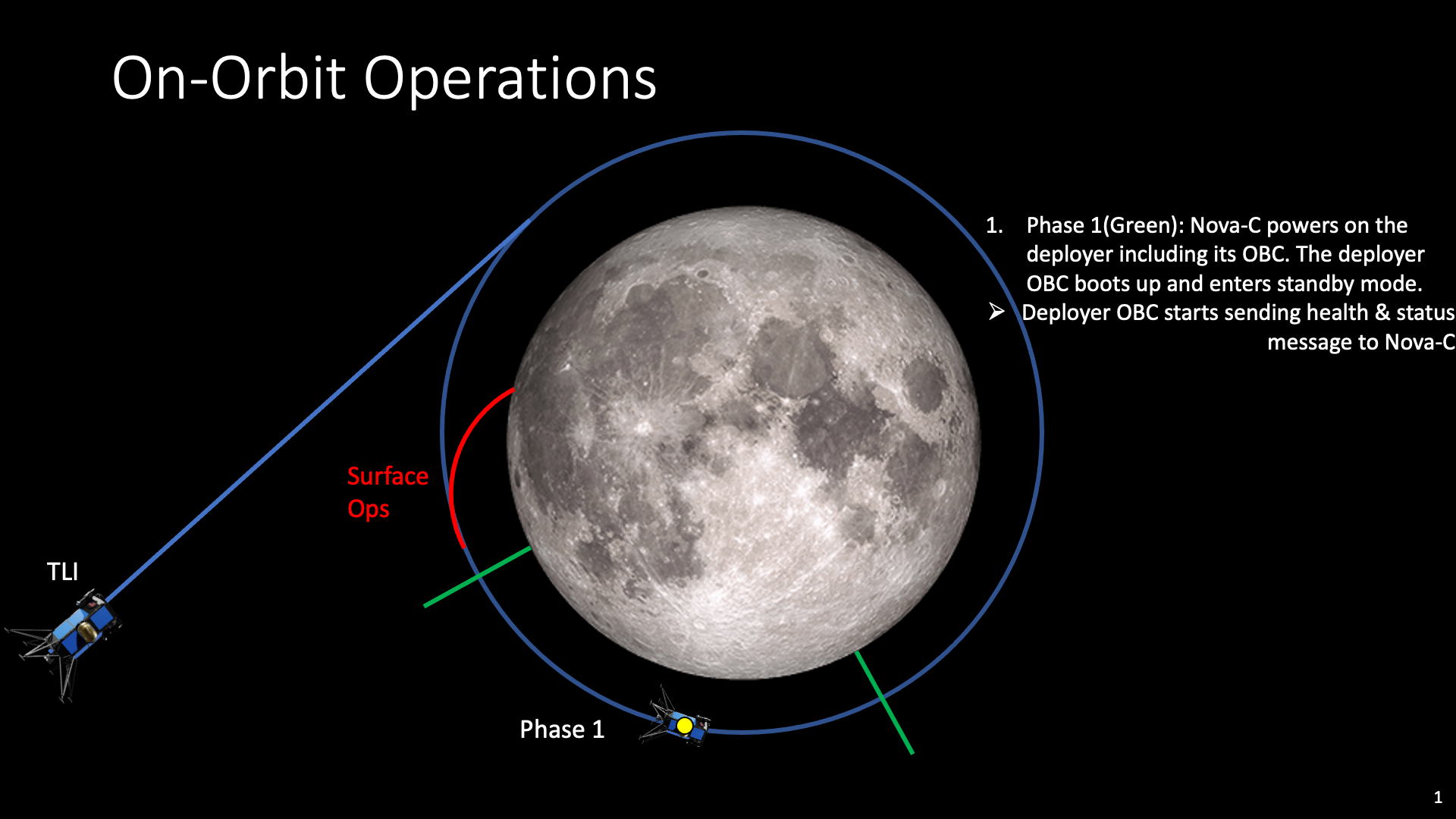}
        \end{minipage}%
        
        \begin{minipage}[]{0.8\textwidth}
            \centering
            \includegraphics[width=\textwidth]{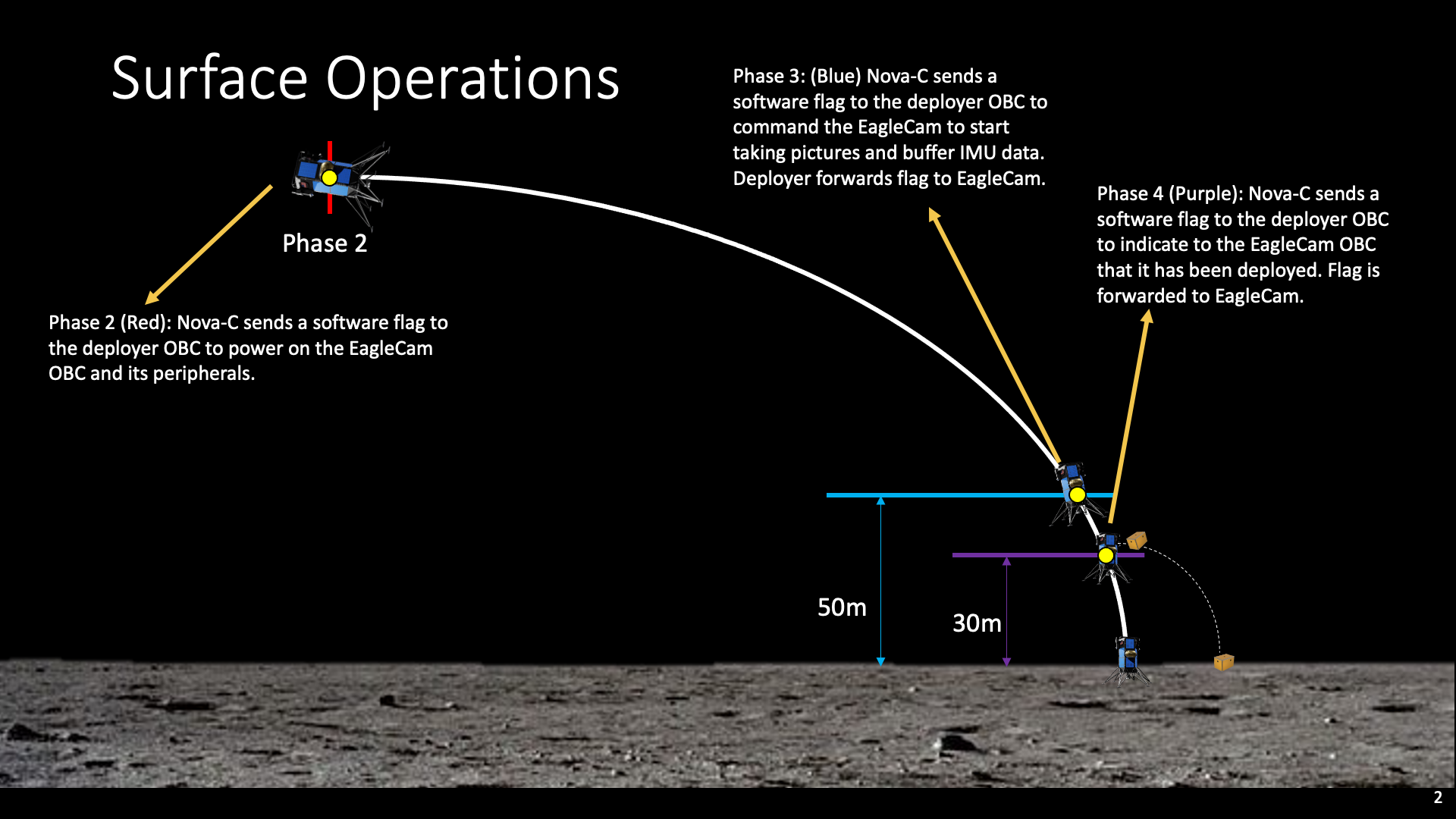}
        \end{minipage}
        \caption{Top: EagleCam's on-orbit ConOps. Bottom: EagleCam's surface ConOps.}
        \label{fig:cfs:dynamics}
    \end{figure}

The EagleCam concept of operations before ejection consists of the following four phases:
\begin{itemize}
    \item Phase 1: Nova-C powers on the deployer including its OBC.
    The deployer OBC boots up and enters standby mode.
    \item Phase 2: Nova-C sends a software flag to the deployer OBC to power on the EagleCam OBC and its peripherals. The deployer OBC reacts to the command and the EagleCam OBC boots up.
    After boot-up, it initializes the peripherals and enters standby mode during which it sends health and status messages to the deployer OBC.
    Then, these messages are forwarded to the Nova-C OBC.
    Then, Nova-C relays these messages to mission control.
    \item Phase 3: At about 50 $m$ altitude above the lunar surface, Nova-C sends a software flag to the deployer OBC to command the EagleCam to start taking pictures and buffer IMU data.
    The deployer OBC forwards this flag to the EagleCam OBC which reacts to it by commanding the cameras' OBCs to start their software that will capture the images.
    \item Phase 4: At 30 $m$, Nova-C sends a software flag to the deployer OBC to indicate to the EagleCam OBC that it has been deployed.
    The deployer OBC forwards this flag to the EagleCam OBC to coordinate data collection and peripheral control. Then, Nova-C triggers a hardware signal to the deployer and deploys EagleCam.
    This phase takes place when Nova-C is performing its vertical terminal descent into the surface as staged in Fig. \ref{fig:cfs:dynamics}.
\end{itemize}

After the ejection maneuver, EagleCam will approach the lunar surface following a ballistic trajectory.
During this maneuver, it will buffer IMU data, capture and send images of Nova-C, in addition to health and status messages to the deployer OBC.
Upon EagleCam landing, which happens approximately six seconds after deployment, EagleCam starts surface operations, which is to capture images with the ArduCam cameras before and after the EDS is activated, dump IMU buffer to storage, complete image download from the NISA OBCs and complete peripheral data transfer to the deployer OBC.
Rsync is being used for file transfer and it runs periodically to sync data between the EagleCam OBC and deployer OBC starting from EagleCam OBC boot up.

Once the lunar regolith settles from the lunar lander landing, the Electrostatic Dust Shield system will be tested. The purpose of the EDS is to clean the lunar dust off the camera lenses.
This technology is a demonstration for the Electrostatics and Surface Physics Laboratory in the Swamp Works at NASA's Kennedy Space Center (KSC).\cite{calle2011active}
If this technology if successful in the lunar environment, it could have multiple applications, particularly to protect hardware such as tools or experiments and astronauts from lunar dust.\cite{gaier2005effects}

The EagleCam mission success criterion
is to transmit at least one picture of Nova-C as it is landing on the lunar surface.
Therefore, if for any reason EagleCam breaks on impact, the mission will still be considered successful as long as the images of the Nova-C landing are transmitted to the deployer OBC in time.
Therefore, the flight software is setup in a way such that it will transmit images from the EagleCam OBC to the deployer OBC as soon as they become available.

\begin{figure}[h!]
    \centering
    \includegraphics[width=0.55\textwidth]{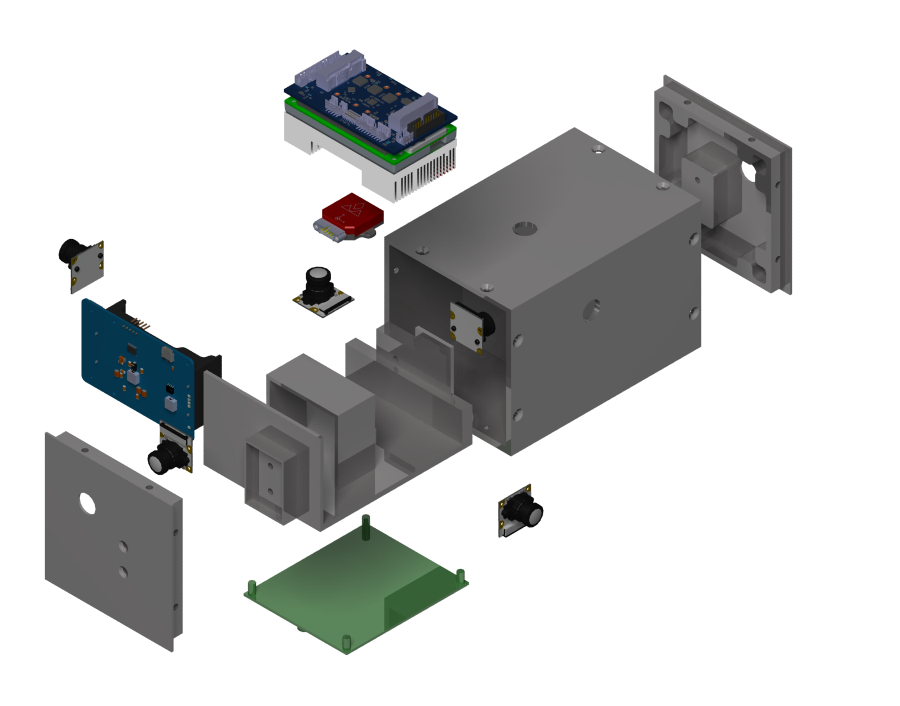}
    \caption{Exploded view of the first iteration of the CubeSat and its components, including the OBC.}
    \label{fig:cfs:architecture}
\end{figure}

EagleCam will experience a variety of thermal environments from being in the shadow during trans lunar injection to being in the sun after deployment.
The lunar thermal environment is based on radiation from the sun and lunar surface, and conduction of the surface.\cite{hurley2015analytic}.
This is important as temperature can physically affect the hardware. During surface operations, the thermal environment of EagleCam is dominated by the radiation and conduction of the lunar surface, which are highly dependent on the sun angle and optical properties \cite{nasa_nedd}.
This is studied and taken into account to guarantee that the software and hardware can operate nominally before the environment surpasses the survivability temperature of the components.\cite{sahr2021wireless} 
The temperature is expected to fall within the range of 262 to 312$\,K$ or -11 to $+39\degree\,C$.
Furthermore, the lunar atmosphere presents a significantly low pressure of $10^{-4}\,Pa$ \cite{finckenor2017researcher}.
This implies that the software must run efficiently to guarantee mission success under this hazardous environment.

\section{Core Flight System Architecture}
cFS is a platform and project independent reusable software architecture, and set of reusable software applications \cite{mccomas2012nasa} with NASA's Technology readiness Level 8-9\cite{cannon2015flight,prokop2018core,cudmore2017porting,cannon2017application}. Therefore, it was chosen for this mission because it provides the base tools for building a modular, reliable, robust, and reusable flight software and because of ease of interface with the Nova-C lander. 


The cFS architecture allows dividing the flight software into separate applications where each application handles a certain function of the mission.
An application can publish its data to other apps to consume, and it can subscribe to data from other apps.
Message routing between apps is handled by the software bus core service.
cFS provides other core services, these are shown in green in Fig. \ref{fig:cfs:eaglecam:architecture} and Fig. \ref{fig:cfs:deployer:architecture}.
The Scheduler app (SCH) which is shown in blue is a cFS app that provides a method of generating software bus messages at pre-determined timing intervals. This allows a system to operate in a Time Division Multiplexed (TDM) fashion with deterministic behavior when combined with a real-time OS such as VxWorks or Preempt RT Linux.


The EagleCam OBC and the deployer OBC are each running an instance of the cFS and both instances are exchanging messages via WiFi.
Each cFS instance is running its own set of applications as depicted in Fig. \ref{fig:cfs:eaglecam:architecture} and Fig. \ref{fig:cfs:deployer:architecture}.

\subsection{EagleCam cFS Implementation}
\begin{figure}[hbt!]
    \centering
    \includegraphics[width=0.8\textwidth]{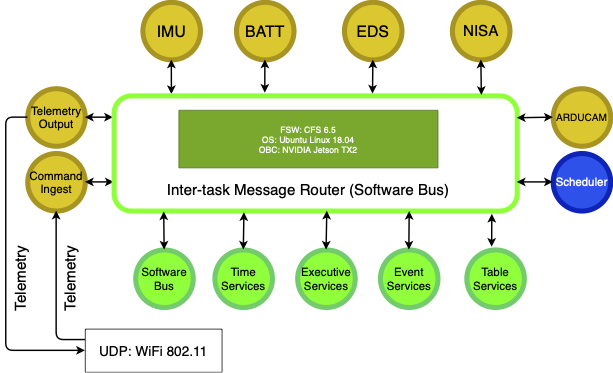}
    \caption{cFS application architecture running on EagleCam's OBC.}
    \label{fig:cfs:eaglecam:architecture}
\end{figure}

\begin{figure}[hbt!]
    \centering
    \includegraphics[width=0.8\textwidth]{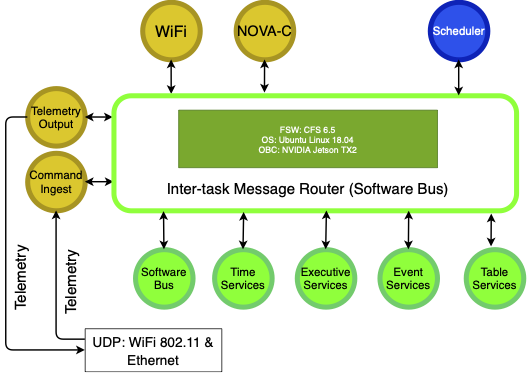}
    \caption{cFS application architecture running on the deployer OBC.}
    \label{fig:cfs:deployer:architecture}
\end{figure}

Fig. 
\ref{fig:cfs:eaglecam:architecture} shows the architecture of the EagleCam cFS instance. Applications were reused from NASA repositories such as Scheduler (SCH), Telemetry Output (TO) and Command Ingest (CI), Other applications were created for the EagleCam mission.
The following list consists of the mission applications that were created which comprise the EagleCam cFS that is running on the EagleCam OBC and their functions:

\begin{itemize}
    \item IMU App: Initializes the IMU and buffers its data before dumping it to a file to be sent to ground control. The app starts buffering data upon receiving mission phase 3 flag from the deployer OBC.
    \item NISA App: Commands the three NISA OBCs to start their software that will take the pictures upon receiving mission phase 3 flag from the deployer OBC.
    \item ArduCam App: Initializes the two ArduCam camera sensors and waits for mission phase flag 3.
    Once the flag has been received, the app waits until EagleCam lands on the lunar surface before it takes a few pictures with the ArduCam cameras, then waits again for the EDS to clean the lenses, then takes a few more pictures after the EDS is done and deactivated.
    \item EDS App: Activates the EDS after a time delay counting from the reception of the mission phase 3 flag, and waits for some time before it deactivates it. The time delay to activate the EDS is programmed to allow enough time for the lunar regolith to settle on the Arducam cameras' lenses. The time during which the EDS is activated is programmed based on experiments conducted to determine how long the EDS takes to clean the lenses.
    \item Battery App: Reads battery status from the battery module and publishes it on the software bus to be sent to ground control as part of the health and status message.
\end{itemize}

\subsection{Deployer cFS Implementation}
Figure \ref{fig:cfs:deployer:architecture} shows the architecture of the deployer cFS instance.
Applications from NASA repositories have been reused such as SCH, TO and CI.
Due to cFS modularity, other applications were created specifically for this mission.
The following list consists of the mission applications that were created which comprise the deployer cFS that is running on the deployer OBC and their functions:

\begin{itemize}
    \item WiFi App: Constantly ensures WiFi connectivity to the EagleCam OBC and checks the status of the deployer WiFi chip.
    The check on the deployer WiFi chip can serve as an indication as to how long the WiFi chip survives the lunar environment after EagleCam has completed its mission.
    \item Nova-C App: Receives mission phase flags from Nova-C and forwards the necessary flags to the EagleCam OBC. It also forwards the health and status message from the EagleCam OBC to Nova-C. 
\end{itemize}

The health and status message is sent as a CCSDS (Consultative Committee for Space Data Systems) consists of the following:
\begin{itemize}
    \item WiFi connection status
    \item Deployer WiFi chip status
    \item EagleCam OBC battery status
    \item EagleCam peripheral data folder size
\end{itemize}

\subsection{Application Architecture}
\begin{figure}[h!]
    \centering
    \includegraphics[width=0.7\textwidth]{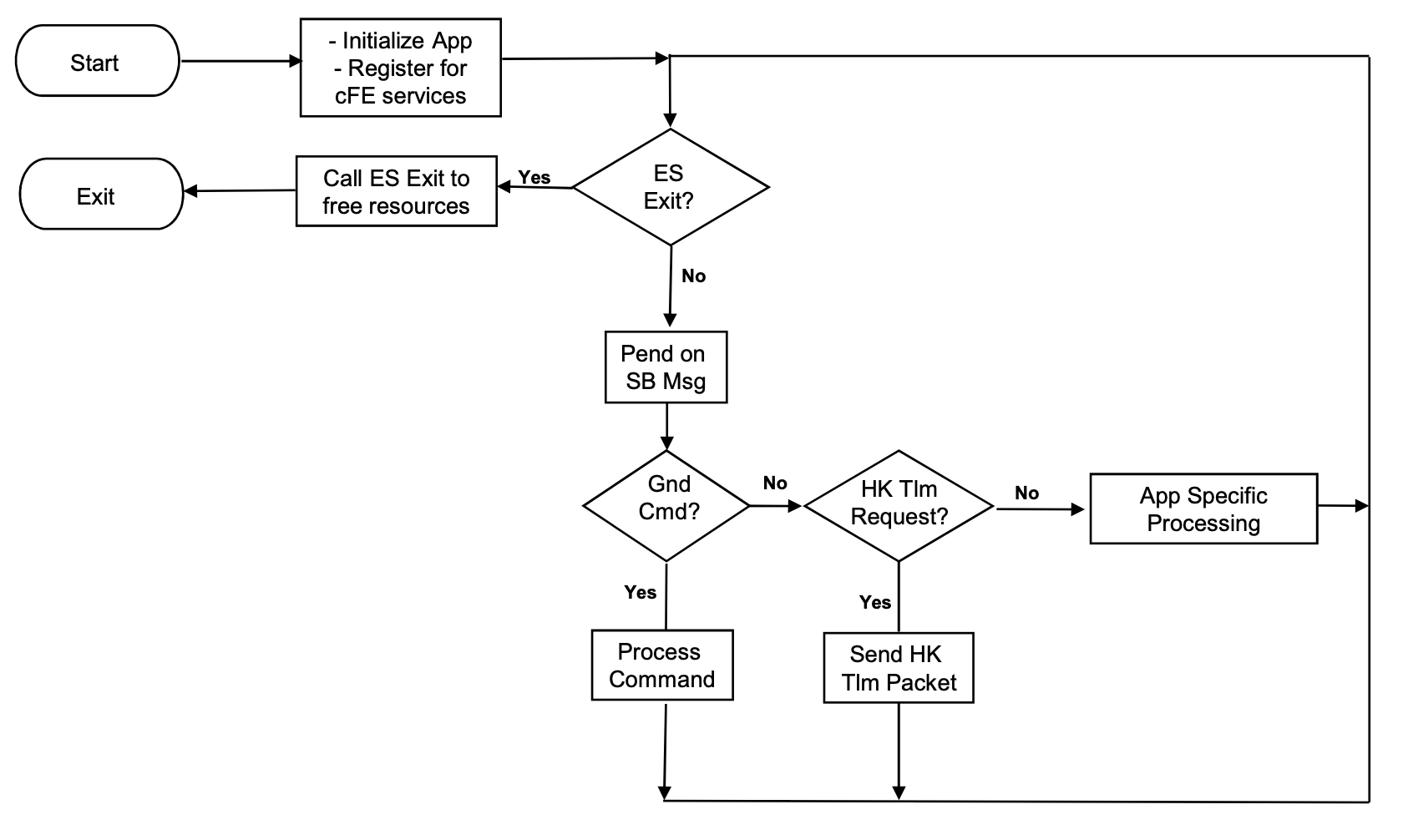}
    \caption{cFS generic app design.}
    \label{fig:generic:app}
\end{figure}

The applications of both subsystems were developed following the general application design\cite{timmons2020core} seen in Fig. \ref{fig:generic:app} suggested by NASA.\cite{mccomas2012nasa}
These applications are programmed in C language.

The complete diagrams for each application can be found in the appendix.

\section{Hardware Implementation}
\begin{figure}[h!]
    \centering
    \includegraphics[width=0.35\textwidth]{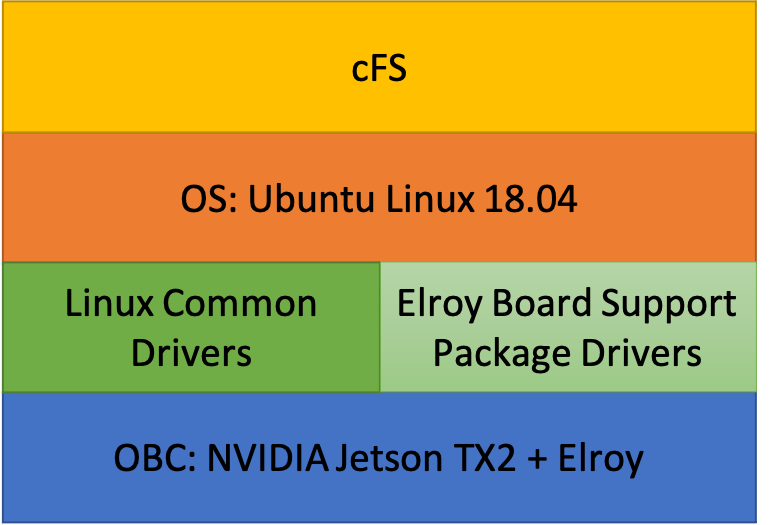}
    \caption{Software layers.}
    \label{fig:cfs:Layers}
\end{figure}

\begin{figure}[hbt!]
    \centering
    \vspace{0.2in}
    \includegraphics[width=0.2\textwidth,trim=8cm 3cm 8cm 4cm]{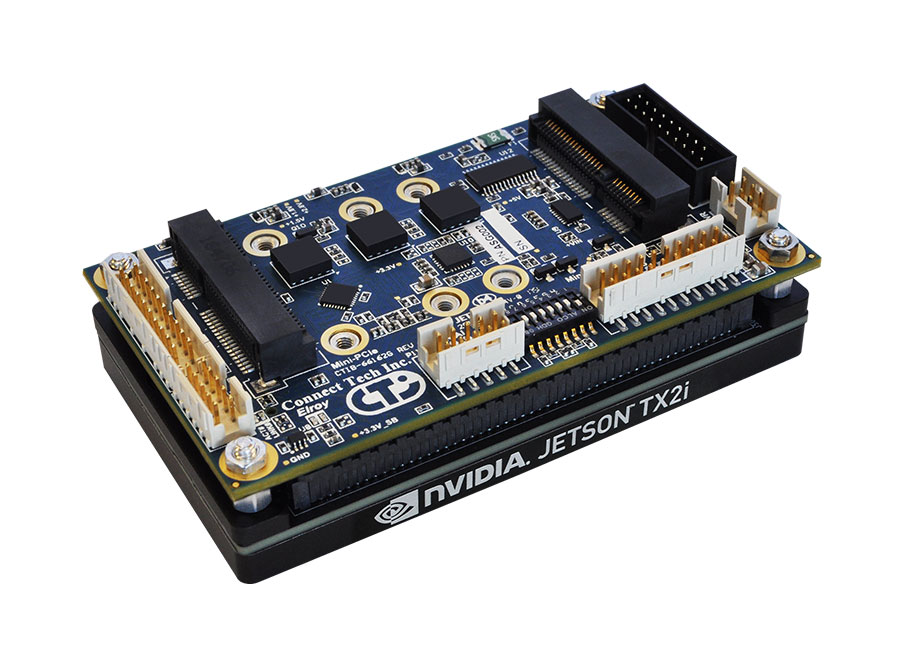}
    \caption{NVIDIA Jetson TX2 with Connect Tech Elroy carrier board.}
    \label{fig:OBC}
\end{figure}

The software was implemented on a NVIDIA Jetson TX2 running Ubuntu Linux 18.04 as seen in Fig. \ref{fig:cfs:Layers}. The Jetson TX2 has one Dual-Core NVIDIA Denver 2 64-Bit CPU and one Quad-Core ARM® Cortex®-A57 MPCore.
The Denver processor is shut down and the software runs completely on the ARM processor.
The TX2 board is connected to a Connect Tech Elroy carrier board (Fig. \ref{fig:OBC}) to provide connections to different peripherals such as USB, Ethernet, and Serial.

\begin{figure}[hbt!]
    \centering
    \includegraphics[width=0.15\textwidth,trim=0.5cm 1cm 0.5cm 2cm, clip]{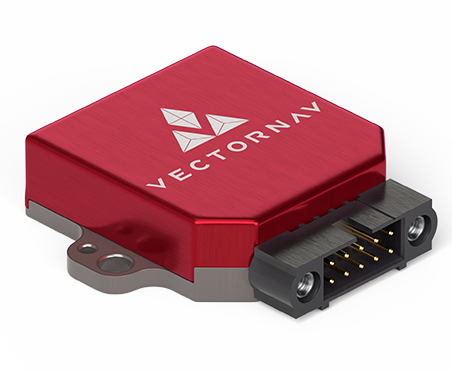}
    \caption{Rugged IMU Vectornav VN-100.}
    \label{fig:IMU}
\end{figure}

To complement the avionics, the Vectornav VN-100 (Fig. \ref{fig:IMU}) IMU is used. This device was configured with a serial data rate of 40Hz to stream and log 3-axis accelerometer filtered and raw data, 3-axis gyroscope data, and 3-axis magnetometer data to try and do a local magnetic field map.

\begin{figure}[hbt!]
    \centering
    \includegraphics[width=0.4\textwidth]{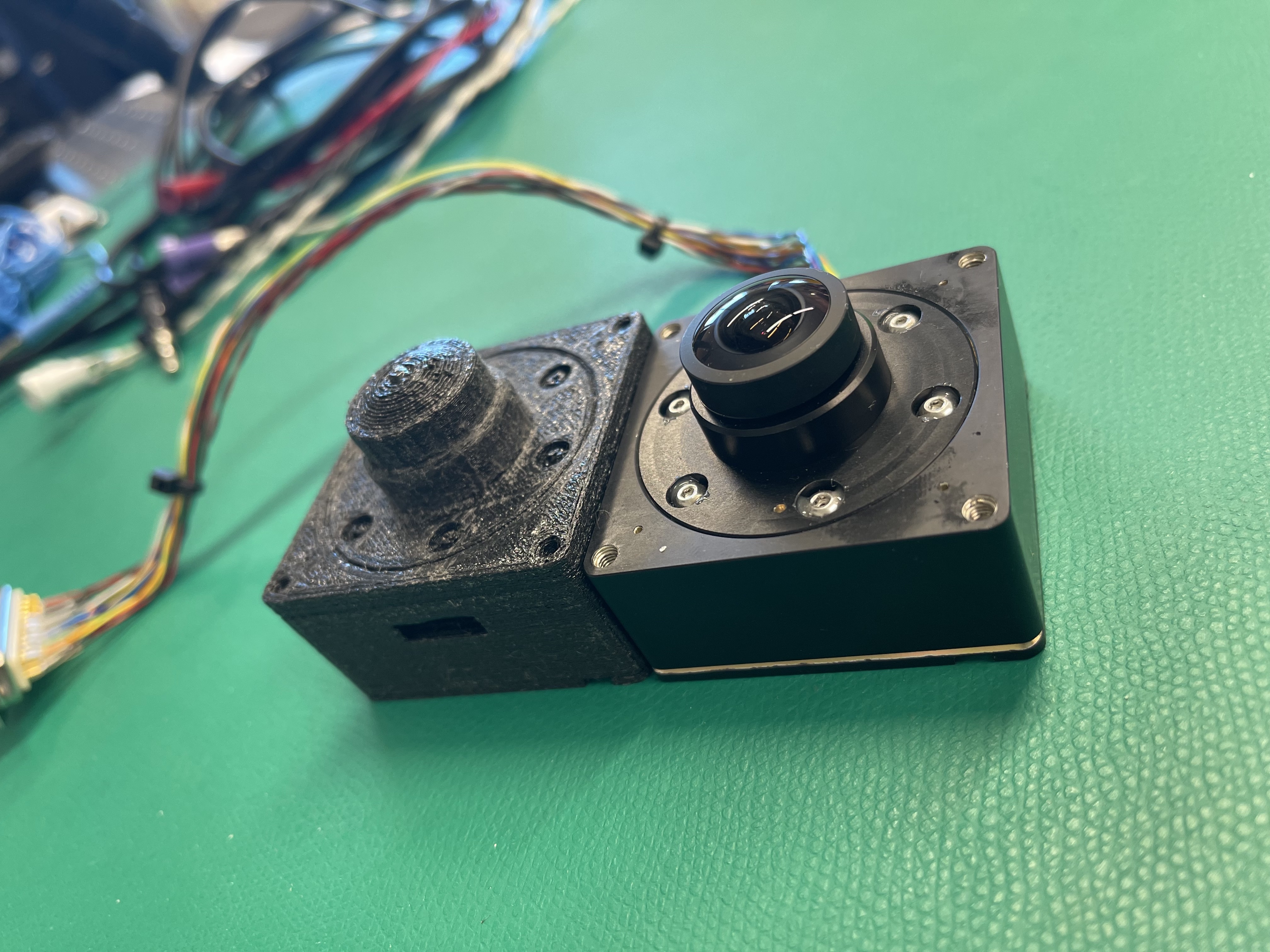}
    \caption{NISA TRL9 Engineering model camera and 3D print of flight qualified hardware.}
    \label{fig:NISA}
\end{figure}

EagleCam will also incorporate three NISA cameras developed by Canadensys which are space-rated (Fig. \ref{fig:NISA}). Each camera is 12MP with a fisheye lens of 186$\degree$ FOV. Therefore, two of these cameras are installed back-to-back on the CubeSat to have a 360$\degree$ FOV. The third NISA camera is located on top of the structure as a redundant sensor to capture the lander, but to also capture images of space.

\begin{figure}[hbt!]
    \centering
    \includegraphics[width=0.75\textwidth]{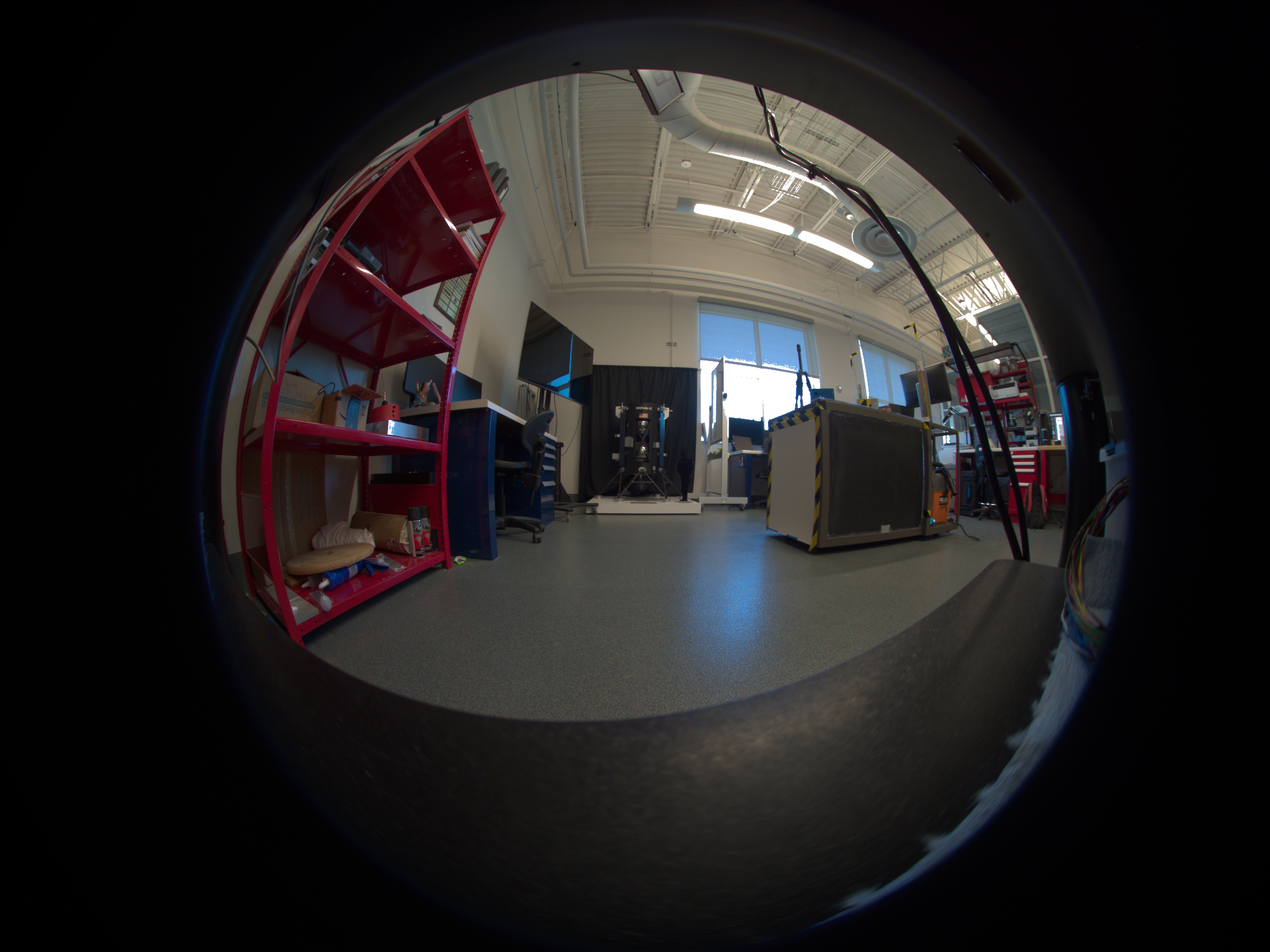}
    \caption{Example of image acquired from the NISA Engineering Model. In the middle of the image a model of the Nova-C lander can be seen. The real lander will be seen approximately in the same proportions as the model that is shown.}
    \label{fig:NISA:example}
\end{figure}

\begin{figure}[hbt!]
    \centering
    \includegraphics[width=0.18\textwidth,trim=6.5cm 10cm 5cm 5cm, clip]{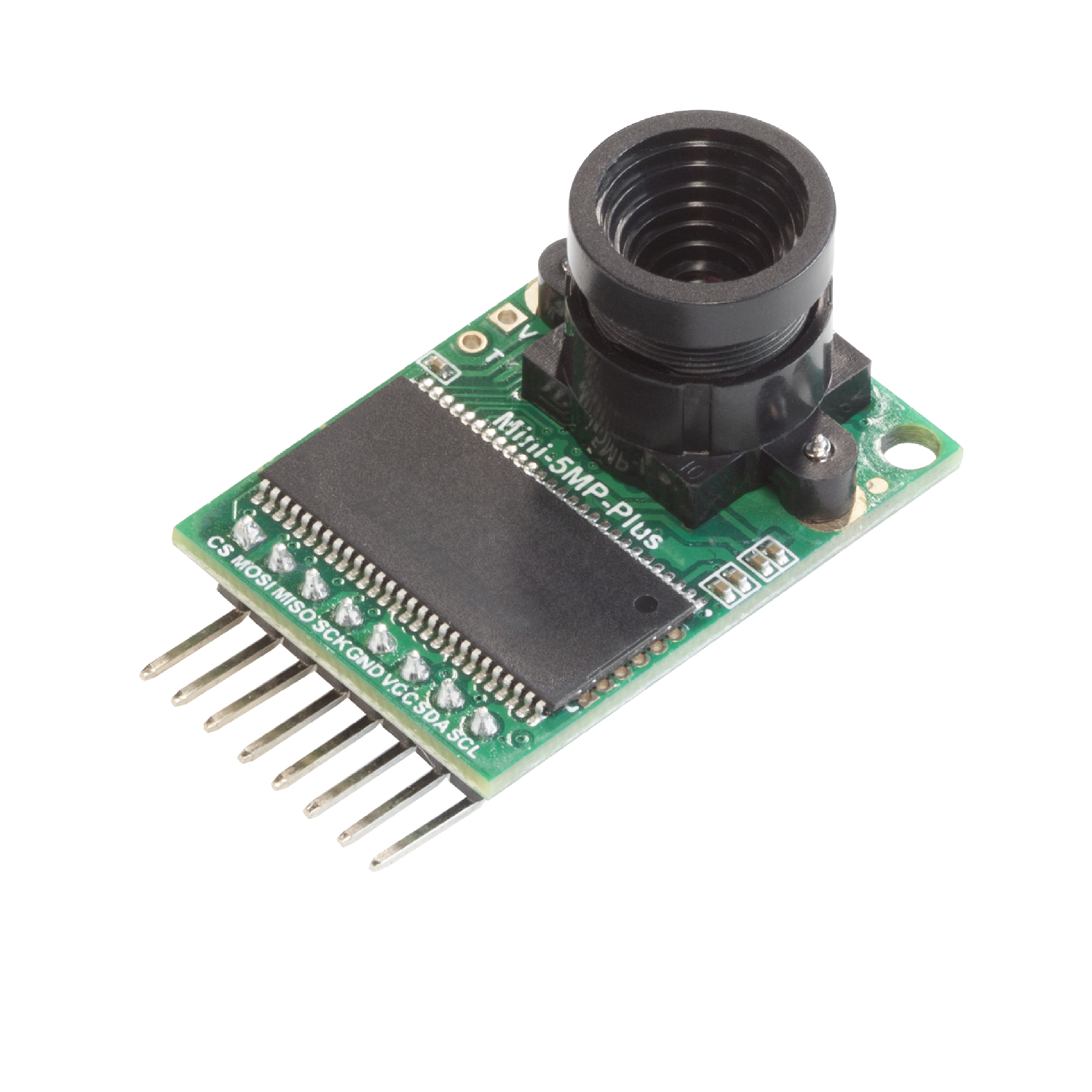}
    \caption{ArduCam with OV5642 sensor.}
    \label{fig:arducam}
\end{figure}

In addition to the NISA cameras, another set of two 5MP camera sensors from ArduCam are used (Fig. \ref{fig:arducam}). These cameras' purpose is to test EDS efficacy and test off-the-shelf components for short-term space missions. One camera will provide redundancy and will serve to provide a different perspective of the lunar landscape. The second camera will be used to perform the technological demonstration of the EDS technology described previously to remove dust from the camera lens. These images can then be used to quantify the cleanliness of the image and how much dust was removed.

\section{System Testing}
The flight software has gone through basic functionality testing and unit testing to ensure stability and reliability. 
In addition to software testing, a significant step for the mission is the testing of the off-the-shelf hardware that are not space-rated to ensure they will contribute to mission success.
Except for the NISA cameras, the Jetson TX2 and its peripherals are not space-rated. Therefore, in-lab tests were performed to ensure they would survive in a space environment.
The hardware was tested for vacuum, harsh temperatures, vibrations, radiation, and impact shock.

For vacuum and harsh temperatures, EagleCam OBC and its peripherals were placed in a thermal vacuum chamber (Fig. 13).
While vacuum was held constant at $10^{-6}\,Pa$, the temperature inside the chamber was dropped to -$45\degree\,C$ and held for 4 hours, then increased to $100\degree\,C$ and held for 4 hours.
Pressure and temperature values were chosen to simulate what EagleCam will experience during its mission.
during this test, the Jetson TX2 was connected to a WiFi network and pinged from a separate laptop to ensure it was still operating inside the chamber.
The peripherals were tested separately afterwards to ensure they survived the test. 
This test will be performed again on the flight unit which will run the final version of the flight software.

    \begin{figure}[h!]
        \centering
        \begin{minipage}[]{0.5\textwidth}
            \centering
            \includegraphics[width=\textwidth]{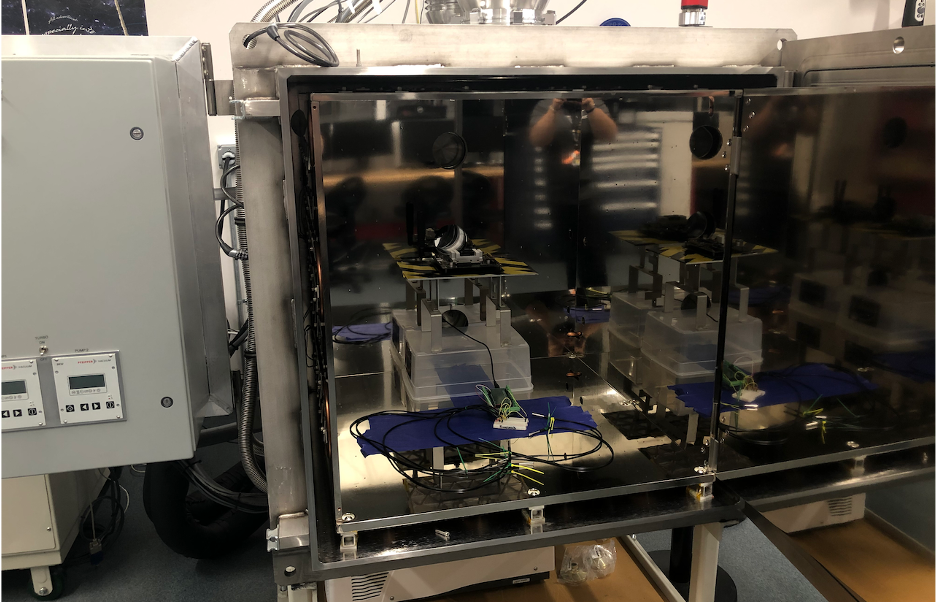}
        \end{minipage}%
        \begin{minipage}[]{0.35\textwidth}
            \centering
            \includegraphics[width=\textwidth]{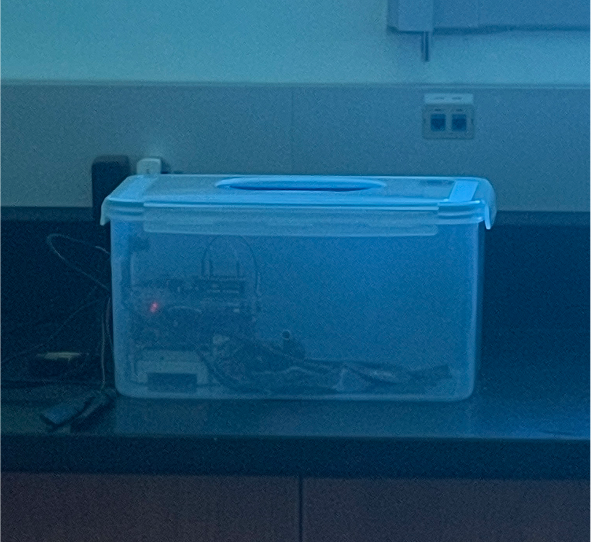}
        \end{minipage}
        \caption{Left: TVAC component and WiFi testing. Right: Radiation testing. }
    \end{figure}

The Jetson TX2 was also tested for radiation (Fig. 13).
It was placed in a container with 3 cesium sources delivering a total amount of 167 $rads$ over a time span of 4 days.
The amount of radiation that EagleCam will experience during its mission lifetime will be less than that experienced in the test. 
In this test, the Jetson TX2 was connected to a WiFi network and running a script to capture images using a webcam.
A laptop was used to ping the Jetson TX2 over WiFi and retrieve the images using FTP.

\begin{figure}[h!]
    \centering
    \includegraphics[width=0.75\textwidth]{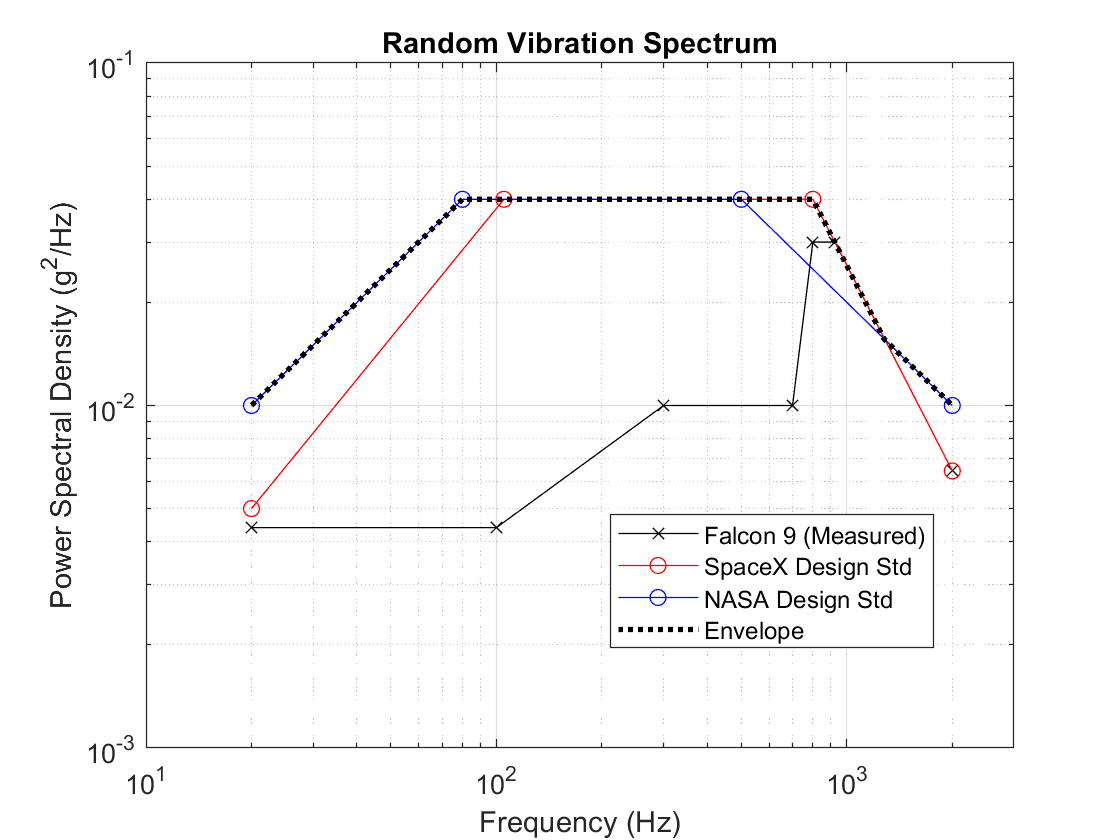}
    \caption{Vibration shaker envelope.}
    \label{fig:honeywell:profile}
\end{figure}

Vibration testing was performed at a Honeywell testing facility.
EagleCam was placed on a shaker and different vibration waveforms were applied on all three axes to simulate the vibrations that will be encountered during launch.
The vibration envelope that was tested is marked with black dashed lines in Fig. \ref{fig:honeywell:profile}.

The Jetson TX2 was turned off during this test since it will be off during launch.
It was powered on afterwards to run the flight software with the IMU, Battery, and WiFi apps.
The test was a success as IMU data was logged, battery status was read, and WiFi connection to a local network was established after power on.

    \begin{figure}[h!]
        \centering
        \begin{minipage}[]{0.5\textwidth}
            \centering
            \includegraphics[width=\textwidth]{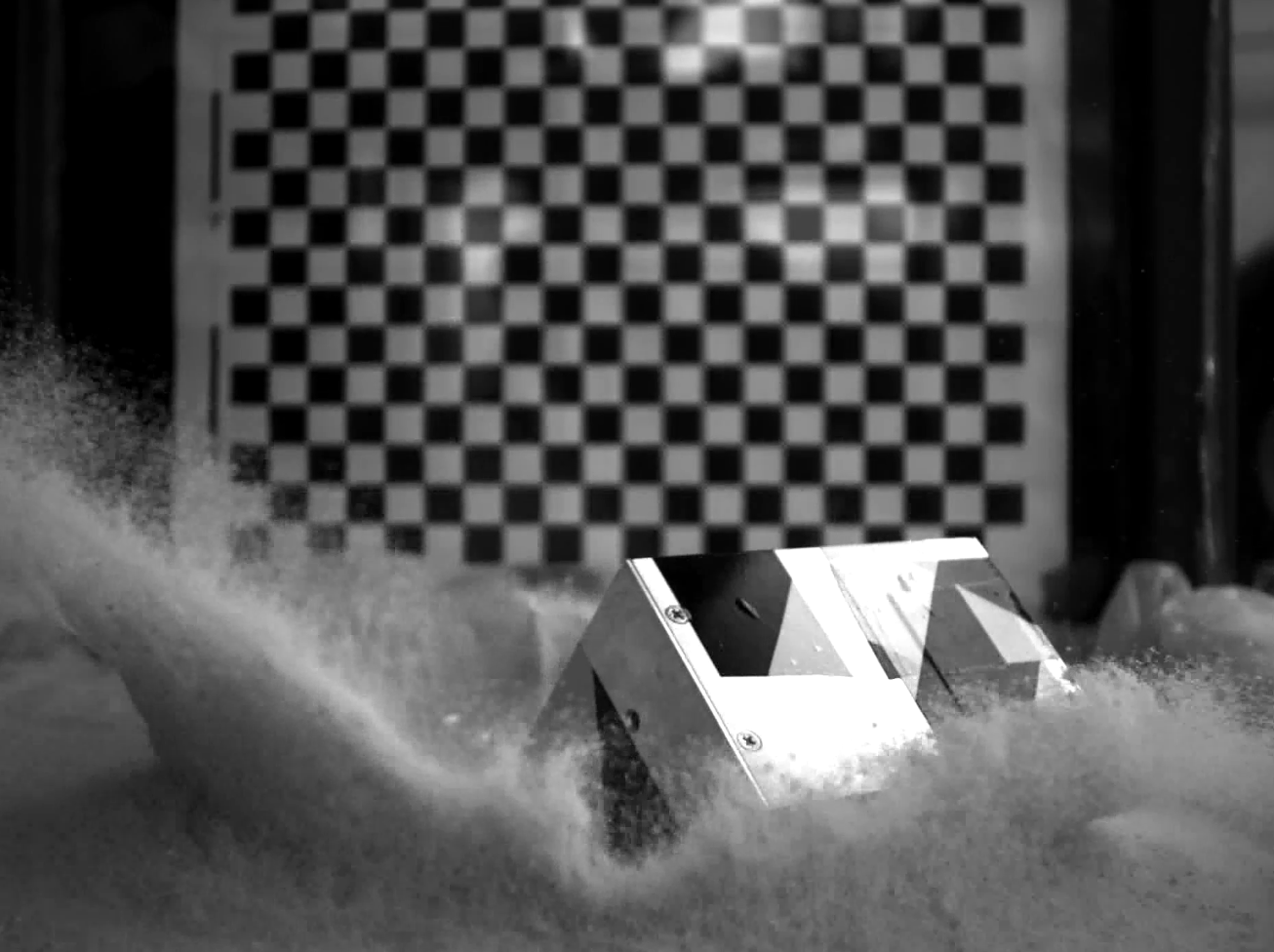}
        \end{minipage}%
        \begin{minipage}[]{0.281\textwidth}
            \centering
            \includegraphics[width=\textwidth]{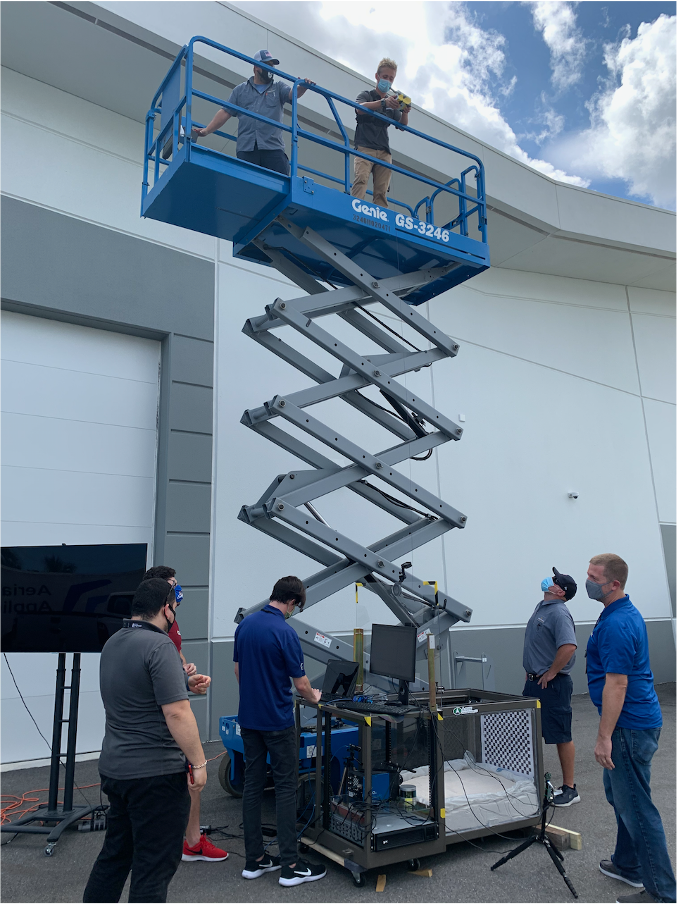}
        \end{minipage}
        \caption{Left: Impact frame. Right: Drop test configuration.}
        \label{fig:droptest}
    \end{figure}

Drop testing was performed to simulate the impact shock EagleCam will experience upon landing on the lunar surface (Fig. \ref{fig:droptest}). The EagleCam structure, which was housing the Jetson TX2, the battery module, and the IMU was dropped from a height of 32 feet $\approx$ 9.7$m$ into a LMS-1 Lunar Mare Simulant container.\cite{hays2021structures}
The IMU application was used in the flight software to log IMU data.
The Jetson TX2 was pinged from a laptop to ensure it was still operating after the impact.
This data was also used to help validate the theoretical impact models and the real impact including the survivability of the structure under shock.


The cFS proves very useful for portability as the apps can be ported easily between architectures making development simpler and faster.



\section{Conclusions}
The cFS software developed by NASA Goddard is a very valuable tool towards promoting a faster software development and implementation for space exploration.
It is flexible and can be used in different platforms.
This software framework is space-rated making it robust and reliable.
This is particularly good given the requirements of EagleCam, providing confidence for mission success.
Different applications can be developed fast using the standard proposed by NASA, allowing easy development of reusable code for multiple hardware architectures.
This was experienced by EagleCam's software developers allowing a quick development of applications to control peripherals and exchange telemetry with the deployer and Nova-C.

\section{Acknowledgments}
The authors would like to gratefully acknowledge Brian Butcher and the team from Intuitive Machines for their help and guidance.

\bibliographystyle{AAS_publication}   
\bibliography{nasacfe}   

\clearpage
\appendix
\section*{Appendix: Applications Flowcharts}
The following diagrams are flowcharts based on the general application design and display how they were adapted to meet the requirements of the mission and provide the appropriate interaction with the different hardware.

\begin{figure}[h!]
    \centering
    \includegraphics[width=\textwidth]{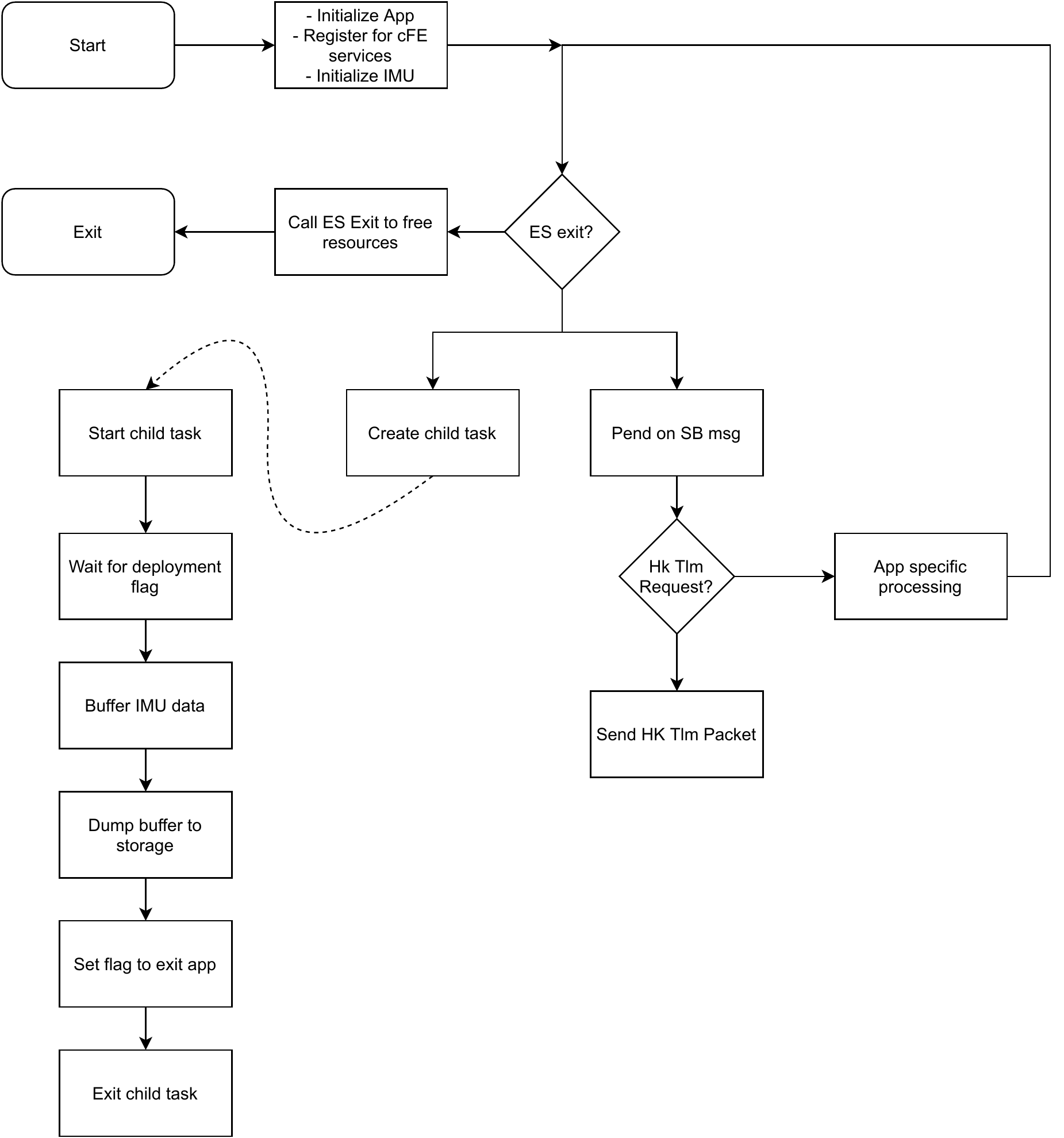}
    \caption{IMU application flowchart.}
    \label{fig:app:IMU}
\end{figure}

\begin{figure}[h!]
    \centering
    \includegraphics[width=0.85\textwidth]{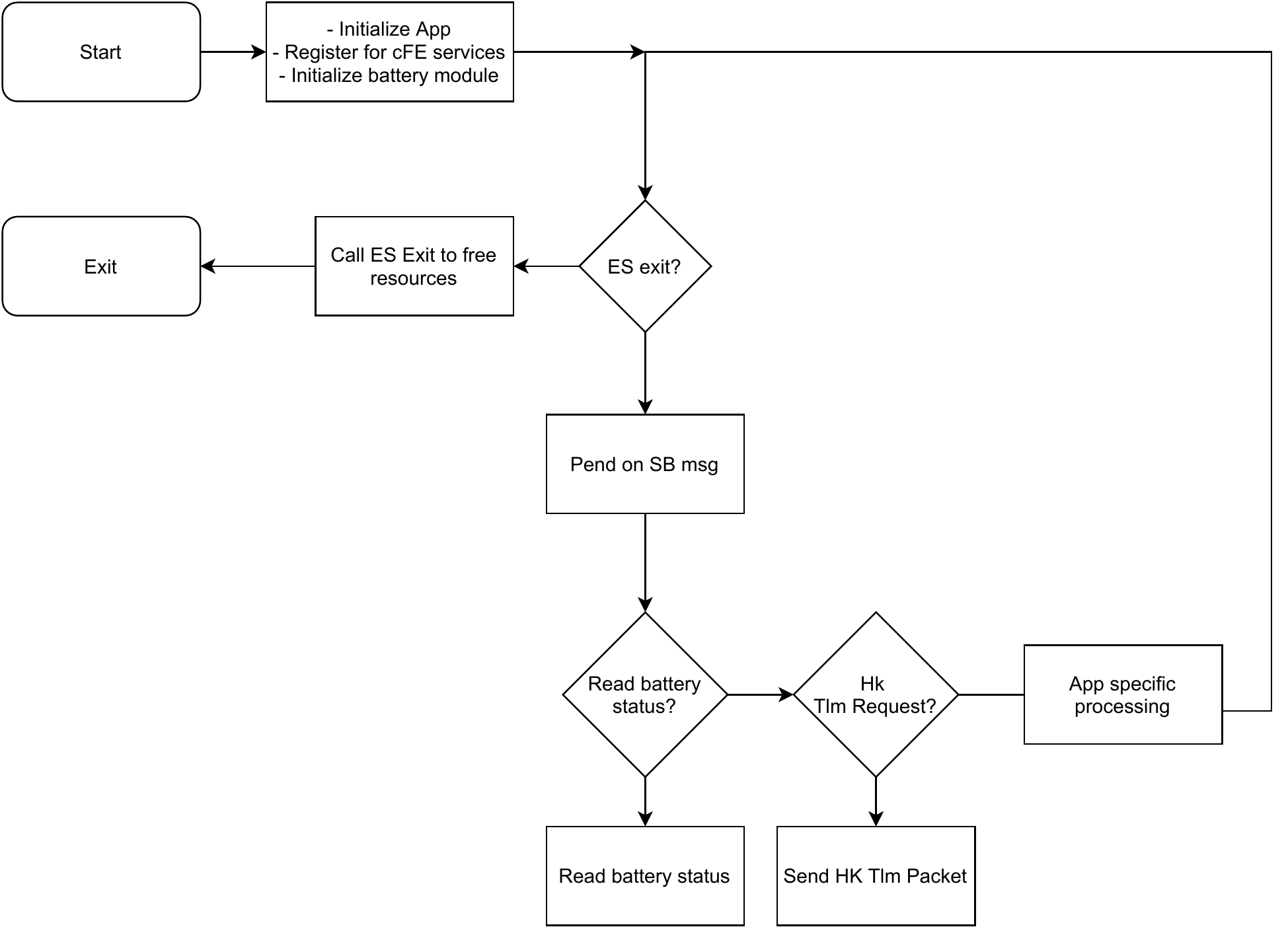}
    \caption{Battery application flowchart.}
    \label{fig:app:PWR}
\end{figure}

\begin{figure}[h!]
    \centering
    \includegraphics[width=0.85\textwidth]{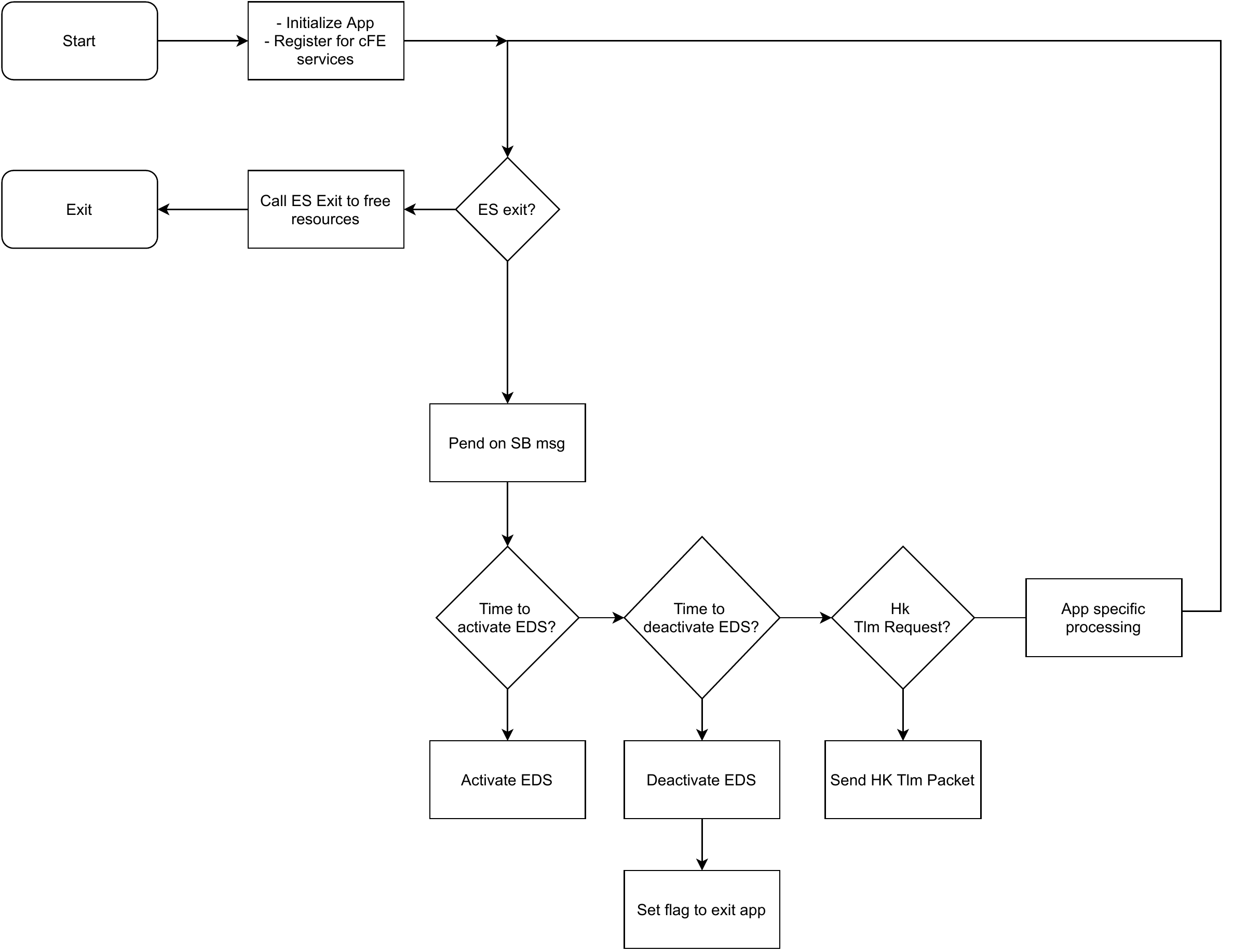}
    \caption{EDS application flowchart.}
    \label{fig:app:EDS}
\end{figure}

\begin{figure}[h!]
    \centering
    \includegraphics[width=0.84\textwidth]{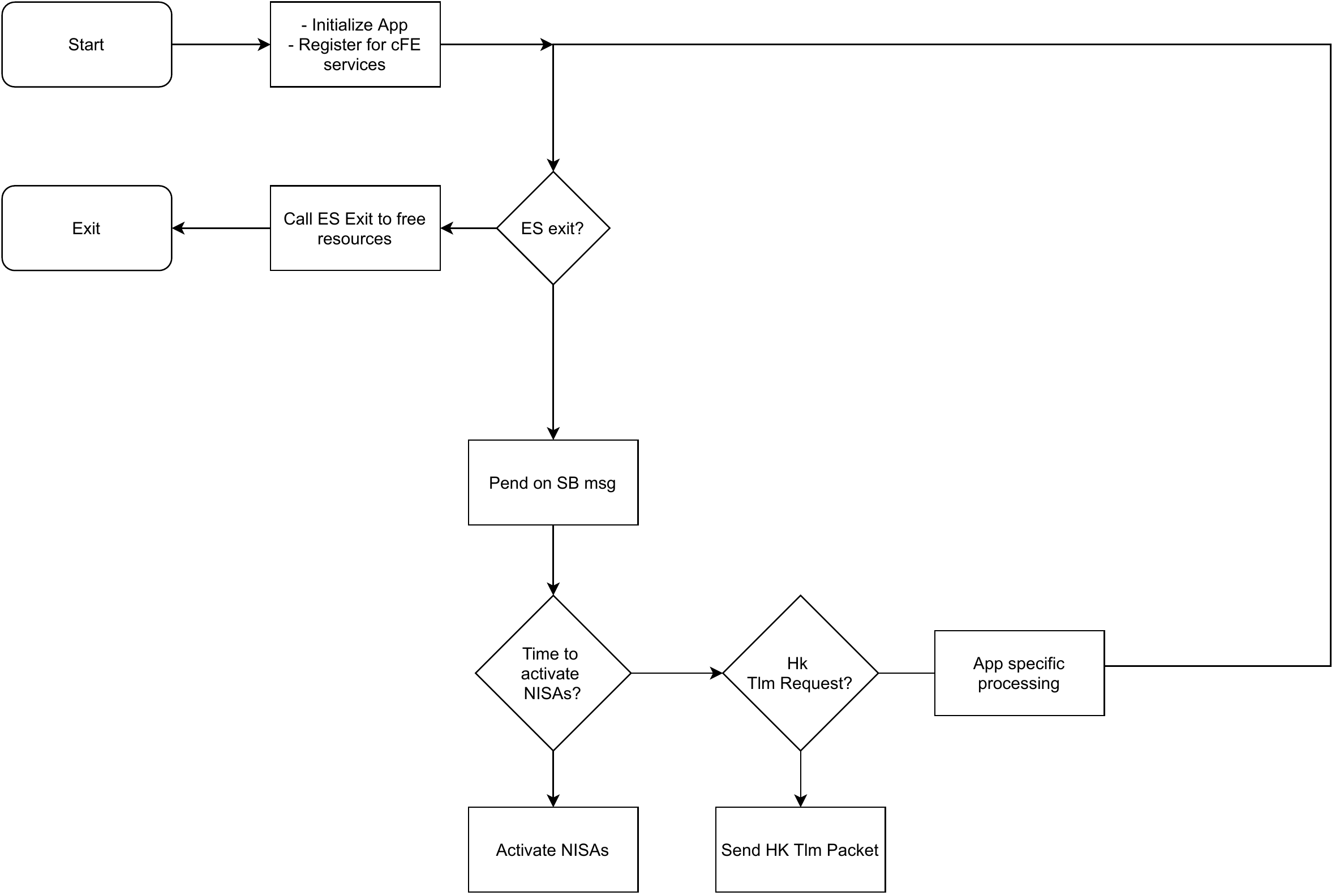}
    \caption{NISA Cameras application flowchart.}
    \label{fig:app:NISA}
\end{figure}

\begin{figure}[h!]
    \centering
    \includegraphics[width=0.86\textwidth]{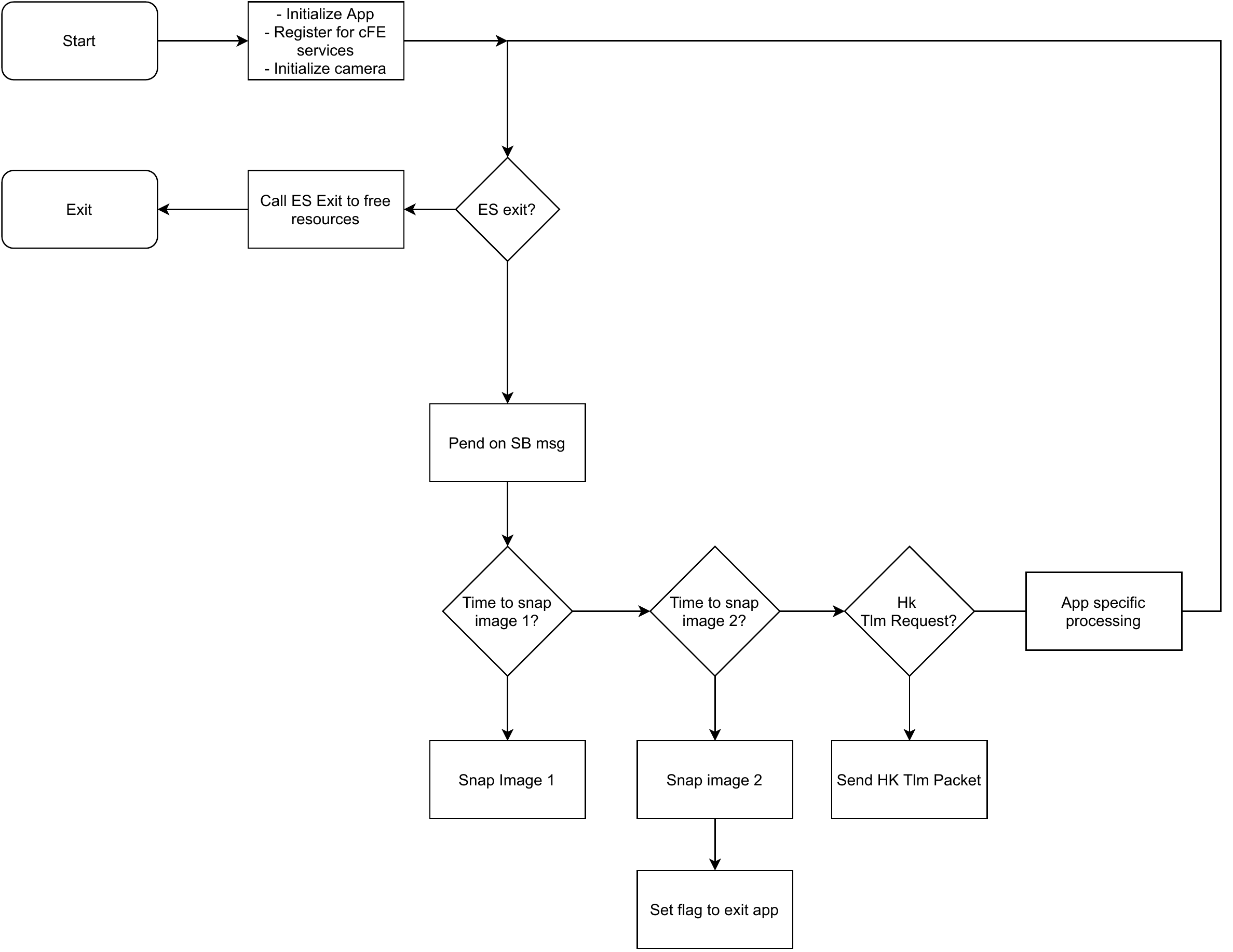}
    \caption{ArduCam application flowchart.}
    \label{fig:app:ARDUCAM}
\end{figure}

\begin{figure}[h!]
    \centering
    \includegraphics[width=0.98\textwidth]{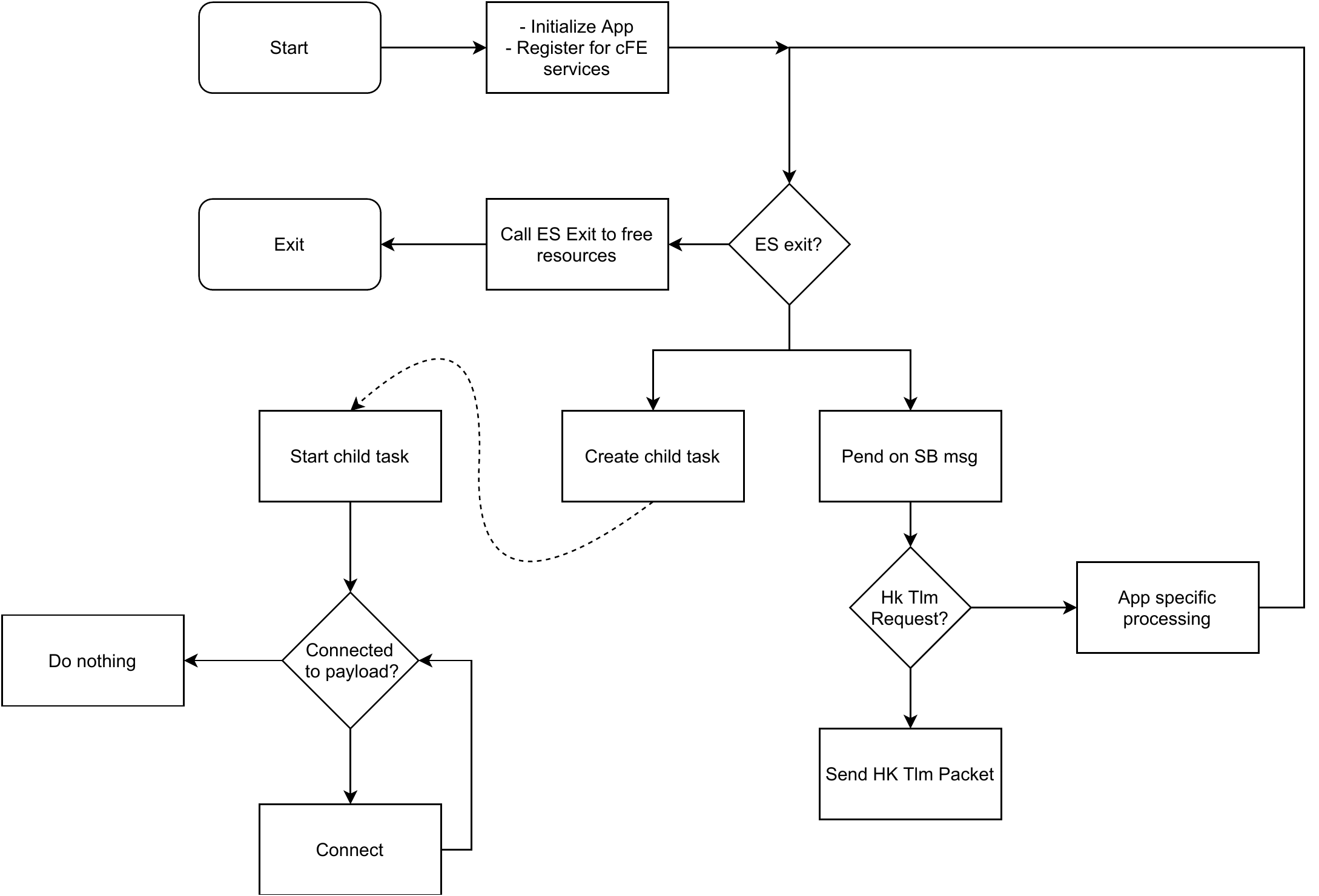}
    \caption{WiFi application flowchart.}
    \label{fig:app:WIFI}
\end{figure}

\begin{figure}[h!]
    \centering
    \includegraphics[width=0.98\textwidth]{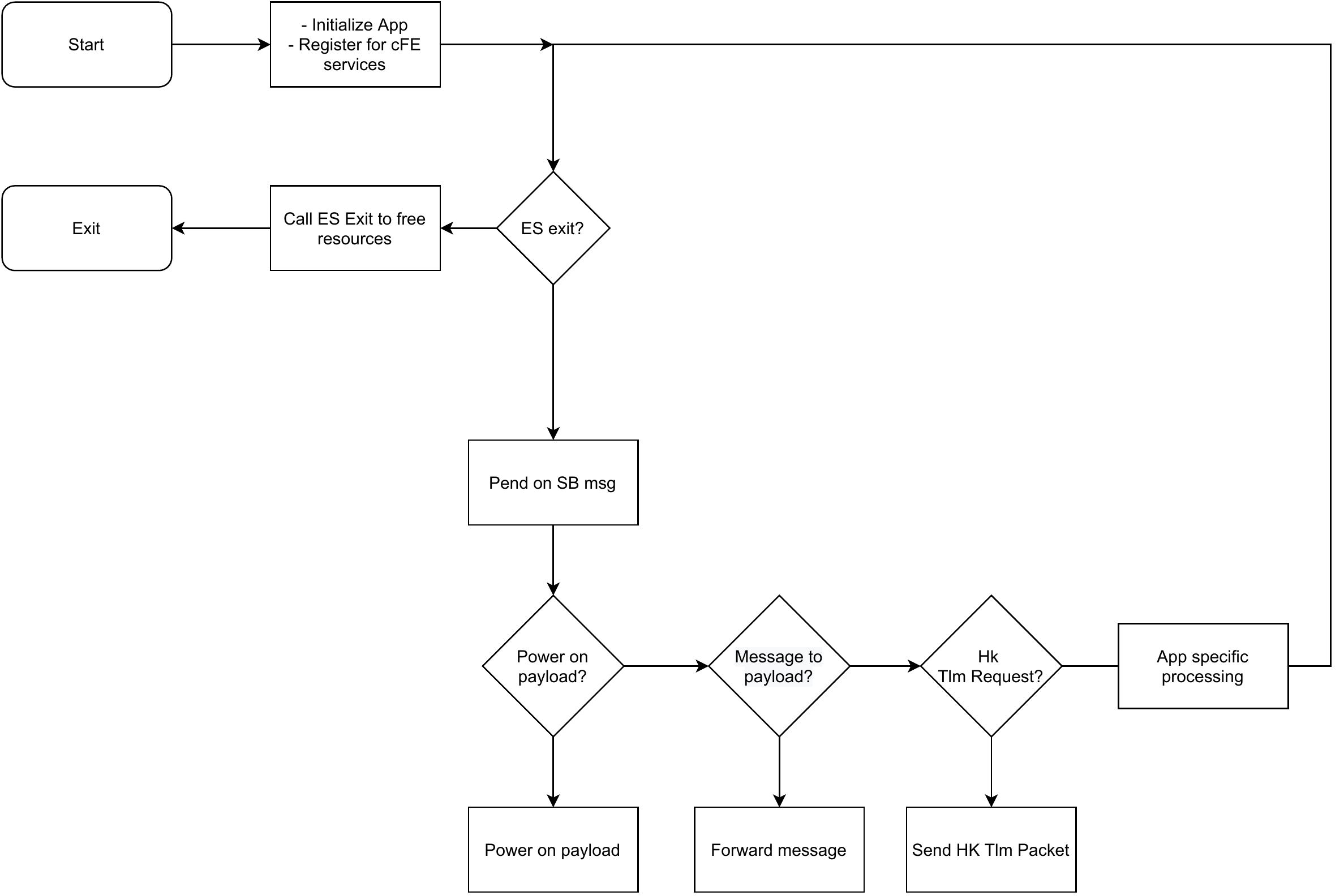}
    \caption{Nova-C application flowchart.}
    \label{fig:app:NVC}
\end{figure}

\end{document}